\documentclass[letterpaper,10pt,conference]{ieeeconf}
\IEEEoverridecommandlockouts % This command is only needed if 
\usepackage{url}
\usepackage{graphicx}
\usepackage{cite}
\makeatletter
\let\NAT@parse\undefined
\makeatother
\usepackage{amsmath}
\usepackage{amssymb}  % assumes amsmath package installed
\usepackage{hyperref}
\usepackage[aboveskip=0.5pt]{subcaption}
\usepackage[skip=0.5pt]{caption}
\usepackage{multirow}
\usepackage{multicol}
\usepackage{booktabs}
\usepackage{color}
\usepackage[ruled,vlined,noend]{algorithm2e}
\providecommand{\keywords}[1]{\textbf{\textit{Index terms---}} #1}
\begin{document}
% \title{Title of the Paper}
\title{\bf{RUSOpt: Robotic UltraSound Probe Normalization with Bayesian Optimization for In-plane and Out-plane Scanning} \vspace{-0.1cm}}
% \author{Author 1, Author 2, Author 3, Author 4, Author 5, Author 6% <-this % stops a space % stops a space
\author{Deepak Raina$^{12*}$, Abhishek Mathur$^3$, Richard M. Voyles$^2$, Juan Wachs$^2$, SH Chandrashekhara$^4$, \\ Subir Kumar Saha$^1$\vspace{-0.0cm}% <-this % stops a space % stops a space
% \thanks{We acknowledge the support for this work from Science and Engineering Research Board (India) and Purdue University (USA) - Overseas Visiting Doctoral Fellowship (Award No. SB/S9/Z-03/2017-VIII); Prime Minster's Research Fellowship (PMRF), IIT Delhi; and Daniel C. Lewis Professorship.}
\thanks{This work was supported in part by SERB (India) - OVDF Award No. SB/S9/Z-03/2017-VIII; PMRF - IIT Delhi under Ref. F.No.35-5/2017-TS.I:PMRF; National Science Foundation (NSF) USA under Grant \#2140612; Daniel C. Lewis Professorship, Amazon Research Award in Robotics and PU-IUPUI Seed Grant.}
\thanks{$^{1}$Indian Institute of Technology (IIT), Delhi, India (\{deepak.raina, saha\}@mech.iitd.ac.in); $^{2}$Purdue University (PU), Indiana, USA (\{draina, rvoyles, jpwachs\}@purdue.edu); $^{3}$IIT, Goa, India (abhishek.mathur.20063@iitgoa.ac.in); $^{4}$All India Institute of Medical Sciences (AIIMS), Delhi, India (drchandruradioaiims@gmail.com).}
\thanks{$^{*}$Corresponding author is Deepak Raina}
}
% IEEE PINS:
% Voyles: 101960; Saha: 102675; Chandru: 306017; Deepak: 226618; Wachs: 146275
% (Ref: F.No.35-5/2017-TS.I: PMRF)
\maketitle
% ############################
% ##### 200 WORDS ############
% ############################
\begin{abstract}
The one of the significant challenges faced by autonomous robotic ultrasound systems is acquiring high-quality images across different patients. The proper orientation of the robotized probe plays a crucial role in governing the quality of ultrasound images. To address this challenge, we propose a sample-efficient method to automatically adjust the orientation of the ultrasound probe normal to the point of contact on the scanning surface, thereby improving the acoustic coupling of the probe and resulting image quality. Our method utilizes Bayesian Optimization (BO) based search on the scanning surface to efficiently search for the normalized probe orientation. We formulate a novel objective function for BO that leverages the contact force measurements and underlying mechanics to identify the normal. We further incorporate a regularization scheme in BO to handle the noisy objective function. The performance of the proposed strategy has been assessed through experiments on urinary bladder phantoms. These phantoms included planar, tilted, and rough surfaces, and were examined using both linear and convex probes with varying search space limits. Further, simulation-based studies have been carried out using 3D human mesh models. The results demonstrate that the mean ($\pm$SD) absolute angular error averaged over all phantoms and 3D models is $\boldsymbol{2.4\pm0.7^\circ}$ and $\boldsymbol{2.1\pm1.3^\circ}$, respectively.
\end{abstract}

\section{Introduction}
Ultrasound imaging, also known as sonography, is a widely used medical imaging technique that can provide valuable information about the structure and function of organs and tissues. It is used in a wide range of diagnostic and therapeutic procedures \cite{chan2011basics}. Ultrasound imaging has several advantages over other imaging techniques such as X-rays or CT scans. For example, it does not use ionizing radiation, making it safer for patients and clinicians. It is also non-invasive and relatively inexpensive compared to other imaging techniques. While ultrasound imaging is a valuable tool in modern medicine, there are also some drawbacks to this technique. The quality of the ultrasound images obtained is highly dependent on the skill and experience of the operator \cite{berg2006operator}. This dependence on the operator's skill and expertise can lead to variability in image quality, which can affect the accuracy of the diagnosis. Additionally, access to expert sonographers is limited in rural areas, leading to long wait times for patients needing ultrasound imaging. Patients who need these services may have to travel to urban areas, which can result in increased travel costs and time \cite{carr2021influence, brief2019top}.

To address the limited availability of sonographers in rural areas, several telerobotic or fully autonomous Robotic Ultrasound System (RUS) have been introduced in recent years \cite{raina2021comprehensive, duan20215g, carriere2019admittance, li2022dual, jiang2023application}. The objective of RUS is to acquire the image with best quality, so that it is easy to diagnose and won’t require a trained/skilled radiologist. By using RUS, it is possible to maneuver ultrasound probe with a consistent, predetermined and precise contact force, position and orientation. These are critical parameters that greatly impact the quality of acquired ultrasound imaging \cite{pinto2013sources}. 
\begin{figure}[t]
	\centering
	%trim={L,B,R,T}
	\includegraphics[trim=0.25cm 8.3cm 4.5cm 2.3cm,clip,width=\linewidth]{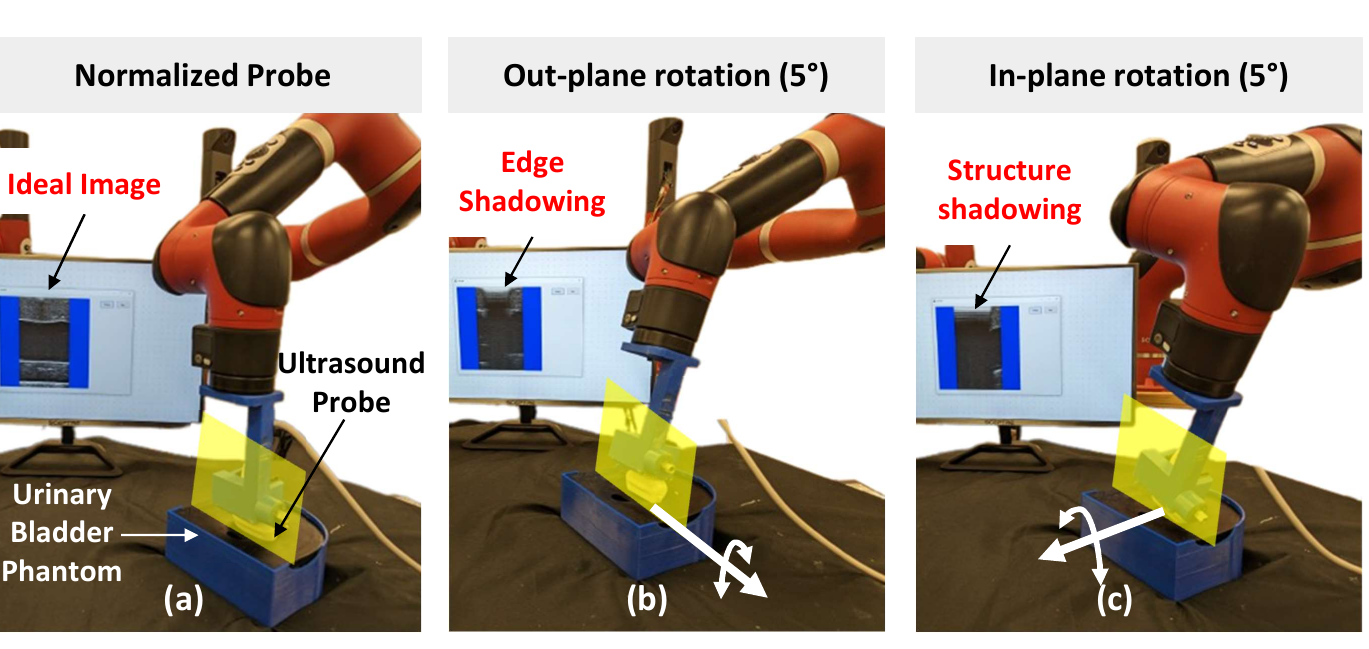}
	\caption{The two rotations carried out to acquire urinary bladder ultrasound images. (a) describe the ideal probe orientation normal to the point of contact on phantom; (b) and (c) describe the non-ideal probe orientation during out-plane an in-plane rotation of probe. The ultrasound images acquired in these cases result in artifacts like edge or structure shadowing due to inappropriate acoustic coupling.}
	\label{fig:teaser}
\end{figure}

In particular, maintaining an appropriate probe orientation is crucial for ensuring the optimal acoustic coupling between the probe and physiologically different human bodies. Fig. \ref{fig:teaser} demonstrates the impact of ideal and non-ideal probe orientation on ultrasound image quality. When the angle between the ultrasound beam and phantom surface normal is zero, more echoes are reflected back and received, resulting in good image quality. However, when the angle is non-zero, resulting ultrasound images present artifacts. Several studies have analyzed the effect of probe orientation on image quality \cite{ihnatsenka2010ultrasound, hoskins2019diagnostic}, indicating that higher quality results can be achieved when the center-line of the probe is aligned with the normal direction of the scanning surface. In orthopedic applications, this phenomenon is even more significant because the intensity reflection coefficient for most soft tissue-to-bone interfaces is approximately $25\%$, compared to less than $0.01\%$ for most soft tissue-to-soft tissue interfaces \cite{hoskins2019diagnostic}. Therefore, it seems reasonable to orient the ultrasound probe along the normal direction of the scanning surface for several ultrasound procedures \cite{kojcev2017reproducibility, huang2018robotic, yang2021automatic}. However, acquiring the normalization of robotized probes remains a challenging task and continues to be of interest to researchers.

% In the field of robotic ultrasound imaging, there have been numerous studies aimed at improving the accuracy and efficiency of probe positioning. One area of particular interest is the development of methods for automatic probe normalization, which involves aligning the ultrasound probe perpendicular to the surface of the tissue being imaged. This is crucial for achieving high-quality ultrasound images and reducing operator variability.

% Despite these limitations, the development of methods for automatic probe normalization represents an important step towards improving the accuracy and efficiency of robotic ultrasound imaging. Continued research in this area could lead to more effective and reliable robotic ultrasound systems, ultimately benefiting both patients and medical professionals.

% The use of robotic ultrasound systems has the potential to revolutionize the way ultrasound imaging is performed. Robotic systems can help reduce operator variability and increase the efficiency of ultrasound imaging, making it easier for medical professionals to obtain high-quality images. However, to realize the full potential of robotic ultrasound systems, accurate and efficient probe positioning is crucial.
\subsection{Related Work on Robotized Probe Normalization}
% {\color{red} It should cover one full column}

Several earlier studies have used 3D patient point cloud data to achieve probe normalization in robotic ultrasound systems \cite{chatelain2017confidence,huang2018robotic, yang2021automatic, al2021autonomous, jiang2020automatic, jiang2022precise, ma2022see}. Huang \textit{et al.} \cite{huang2018robotic} approximated the normal direction of the triangle formed by three neighboring points around the intended scanning path using a depth camera image prior to ultrasound scanning. However, the surface acquired before the probe makes contact with the tissue may not adjust to deformations caused by contact. Further, the method's precision is restricted by hand-eye calibration and potential obstructions between the camera and the scanned tissue. Chatelain \textit{et al.} \cite{chatelain2017confidence} proposed a visual servoing method for in-plane orientation adjustment of the probe for 3D ultrasound imaging. However, the method fails to optimize the out-plane orientation, which is controlled using the teleportation of a robotic arm. Jiang \textit{et al.} \cite{jiang2020automaticconfidence} presented a method for optimizing the in-plane orientation of robotic ultrasound probe based on confidence map-based \cite{karamalis2012ultrasound, hung2021ultrasound} assessment of ultrasound image quality. For out-plane orientation, they further used the force measurements to align central axis of probe with the normal to anatomical surface. However the method requires exploring the full fan motion of the probe at each contact point, which would compromise the scanning speed. Another work by Jiang \textit{et al.} \cite{jiang2020automatic} proposes a mechanical model, which used only force sensor data to estimate the surface normal direction. This method has been tested on phantom and in-vivo tissues. However, the results show that the model fails to accurately identify the normal when rotation limits are close to goal orientation or asymmetrical, which violates the underlying assumptions of their mechanical model. The recent work by Jiang \textit{et al.} \cite{jiang2022precise} extended their work to probe normalization in the presence of patient movements. They developed a motion monitoring module to recompute the motion trajectory in case of patient movement. For probe orientation correction, they used the confidence-maps of ultrasound images to differentiate between the well-connected part and the non-contacting part.

There have also been studies that developed hardware systems to achieve probe normalization in robotic ultrasound systems. Ma \textit{et al.} \cite{ma2022see} proposed a robotic ultrasound imaging system that uses laser distance sensors on the end-effector of robotic arm to achieve normal positioning of the probe. However, the end-effector may fail to identify the normal on highly complex surfaces due to larger outer-case for laser sensor mounting. Moreover, the laser sensor performance might get affected due to different lighting conditions \cite{konolige2008low}. Tsumura \textit{et al.} \cite{tsumura2020robotic, tsumura2021tele} introduced a gantry-style robot with a passive 2-DOF end-effector that ensures the correct positioning of the ultrasound probe in real-time for both obstetric and lung examinations. This end-effector permits swift angle adjustments, as there are no delays in mechanical rotational motion. Nonetheless, the passive mechanism of this device necessitates a broad tissue contact area, limiting its flexibility.

\subsection{Contributions}
Despite several previous attempts to solve the challenges in robotized probe normalization, the previous methods have the following two major drawbacks: (i) they rely on computational-intensive algorithms to extract the skin surface from the depth camera, and (ii) they require time-consuming exploration phase with a large number of steps for identifying the normal direction. In this work, we propose an autonomous RUS by addressing the important challenge of identifying the normal direction with respect to the patient body. We propose a sample-efficient machine-learning-based optimization algorithm, Bayesian Optimization (BO), to search for normal direction by just using the force sensor data without requiring the 3D patient information and thoroughly exploring the rotations. Earlier, BO has been employed in various safety-critical robotic medical procedures, including but not limited to autonomous robotic palpation \cite{yan2021fast}, semi-autonomous surgical robot \cite{chen2020supervised}, tuning of hip exoskeletons' controller \cite{ding2018human}, and Autonomous-RUS \cite{goel2022autonomous}. Goel \textit{et al.} \cite{goel2022autonomous} used BO to search for high image quality region, where only 2D probe position is considered with compliant control. However, orientation adjustment of probe is not considered, which is essential to conduct the clinical ultrasound procedure for patients with different physiological conditions. The \textit{key contribution} of our work are as follows:
\begin{enumerate}
    \item We propose a sample-efficient Bayesian Optimization (BO)-based formalism that will identify the necessary orientation adjustment required for  the robotized ultrasound probe to align the probe in a normal direction to the scanning surface.
    \item We propose a objective function for BO that leverages the force sensor data and underlying mechanics to guide the normal identification process. Further, it handles the noise in objective function measurements by incorporating a regularization term.
    \item We validated the proposed method on urinary bladder phantom with planar, tilted and rough scanning surfaces for normalizing both linear and convex probe. Further, we evaluated its robustness by applying it for 3D mesh human models, which simulate the actual physiology of humans.
\end{enumerate}

\begin{figure*}[!ht]
	\centering
	%trim={L,B,R,T}
	% \includegraphics[trim=0cm 3cm 0cm 0cm,clip,width=\linewidth]{figs/overview9}
	\includegraphics[trim=0cm 10cm 0cm 0cm,clip,width=\linewidth]{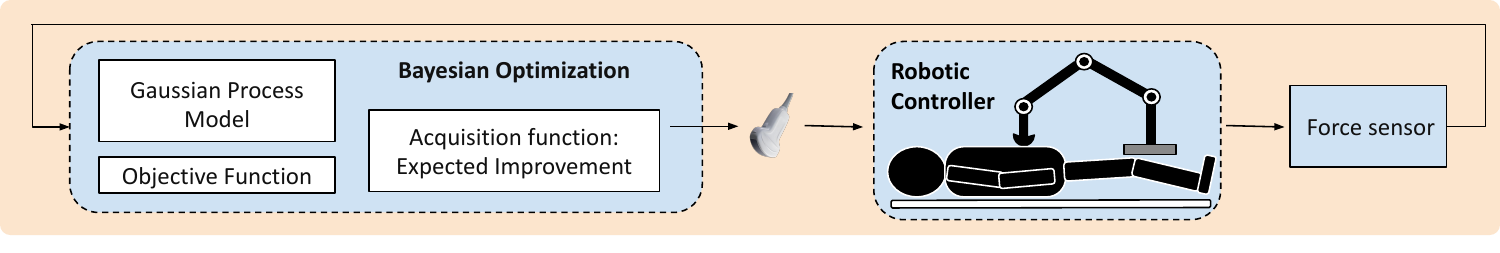}
	\caption{Overview of the BO framework for identifying the normal of the robotized ultrasound probe to scanning region.}
	\label{fig:overview}
\end{figure*}
\section{Methodology}
%{\color{red}{Require flowchart diagram} }
Figure \ref{fig:overview} provides an overview of the proposed methodology, which aims to identify the normal direction with respect to the contact surface during robotic ultrasound procedures. To achieve this, the proposed methodology employs Bayesian Optimization (BO), which uses only force sensor measurements from a wrist-mounted force sensor. An objective function for BO has been formulated using underlying contact mechanics to guide the identification process. BO computes the probabilistic estimate of the unknown \textit{objective function} in the scanning region of the human body and uses an \textit{acquisition function} to select the next best pose to evaluate based on the current belief about the function. Algorithm $1$ provides a detailed flowchart of the proposed methodology.
\subsection{Bayesian Optimization formulation}
If $A$ represents the region of scanning on human body, then the objective function of BO can be formulated as:
\begin{equation}
    \max_{\boldsymbol{p} \in A} {\mathcal{F}}(\boldsymbol{f}(\boldsymbol{p}))
\end{equation}
where $\boldsymbol{f}(\boldsymbol{p}) = [f_x, f_y, f_z]$ represents the probe forces at pose $\boldsymbol{p}$ along the three axis $x$, $y$ and $z$. ${\mathcal{F}}$ represents the unknown \textit{objective function} to be maximized during the BO search. Note that the probe pose represents a vector consisting of probe position ($\boldsymbol{x}$) and orientation ($\boldsymbol{o}$) as $\boldsymbol{p} = (\boldsymbol{x}, \boldsymbol{o})$. The estimator used in BO is Gaussian Process (GP) model, which defines an unknown function ${\mathcal{F}}$ by the mean function $\boldsymbol{\mu}(\cdot)$ and covariance or kernel function $\boldsymbol{\kappa}(\cdot,\cdot)$. If a set of unknown function value estimates is given by $\bar{\boldsymbol{\mathcal{F}}} = [{\mathcal{F}}(\boldsymbol{f}(\boldsymbol{p}_1), \cdots, {\mathcal{F}}(\boldsymbol{f}(\boldsymbol{p}_n)]$ at set of probe poses, given by $\boldsymbol{\bar{p}} = [\boldsymbol{p}_1, \cdots, \boldsymbol{p}_n]$, then the GP regression can predict the unknown function value estimate at new probe pose $\boldsymbol{p}^\dagger$ and is given by
\begin{equation}
    \mathcal{P}({\mathcal{F}}(\boldsymbol{f}(\boldsymbol{p}^\dagger)|\boldsymbol{p}^\dagger,\bar{\boldsymbol{p}}, \bar{\boldsymbol{\mathcal{F}}}) = \mathcal{N}(\boldsymbol{k} \boldsymbol{K}^{-1} \bar{\boldsymbol{\mathcal{F}}}, \boldsymbol{\kappa}(\boldsymbol{p}^\dagger, \boldsymbol{p}^\dagger) - \boldsymbol{k} \boldsymbol{K}^{-1} \boldsymbol{k}^T)
\end{equation}
where,
\begin{equation*}
\boldsymbol{k} = \begin{bmatrix}
\boldsymbol{\kappa}(\boldsymbol{p}_*, \boldsymbol{p}_1) & \cdots & \boldsymbol{\kappa}(\boldsymbol{p}_*, \boldsymbol{p}_n)
\end{bmatrix} 
\end{equation*}
\begin{equation*}
    \boldsymbol{K} = \begin{bmatrix}
\boldsymbol{\kappa}(\boldsymbol{p}_1, \boldsymbol{p}_1) & \cdots & \boldsymbol{\kappa}(\boldsymbol{p}_1, \boldsymbol{p}_n)\\
\vdots & \ddots & \vdots \\
\boldsymbol{\kappa}(\boldsymbol{p}_n, \boldsymbol{p}_1) & \cdots & \boldsymbol{\kappa}(\boldsymbol{p}_n, \boldsymbol{p}_n)
\end{bmatrix}
\end{equation*}
In the above formulation, the kernel matrix $\boldsymbol{\kappa}$ has been formulated using the sum of radial basis function and white kernel function as:
\begin{equation} \label{eq:kernel}
   {\kappa}(\boldsymbol{p}_i, \boldsymbol{p}_j) = \sigma_r \exp \bigg(\frac{-||\boldsymbol{p}_i - \boldsymbol{p}_j||^2}{2l^2}\bigg) + \sigma_w \boldsymbol{{I}_n}
\end{equation}
where $\sigma_r$, $\sigma_w$ and $l$ is the overall variance, noise variance and length-scale, respectively, representing the BO hyperparameters $\boldsymbol{\theta}$. To optimize the performance of a GP model with the kernel in eq. \eqref{eq:kernel}, the hyper-parameters represented by $\boldsymbol{\theta}$ need to be estimated. This is done by maximizing the log marginal likelihood using a Limited memory Broyden–Fletcher–Goldfarb–Shanno (L-BFGS) solver, which is an algorithm used for numerical optimization. 
\begin{equation}
    \boldsymbol{\theta}^* = \arg\max_{\boldsymbol{\theta}} \log \prod \mathcal{N} ({\mathcal{F}}(\boldsymbol{f}(\boldsymbol{p}_i))| \boldsymbol{\mu}_{\boldsymbol{\theta}}(\boldsymbol{p}_i), \boldsymbol{K})
\end{equation}
\begin{algorithm}[h]
\SetAlgoLined
\textbf{Input:} Scanning Region $A$, max. iterations $N_{max}$\;
Initialize $\Bar{\mathbf{p}} = \{\}$, $\Bar{\boldsymbol{\mathcal{F}}} = \{\}$\;
  \For{$i = 1,...,N_{max}$ }{
    $\boldsymbol{p}_i \leftarrow \text{argmax}_{\boldsymbol{p} \in A} EI(\boldsymbol{p})$\;
    \eIf{termination criteria reached}{
        stop\;
    }{
        Probe at~$\boldsymbol{p}_i$, compute objective function ${\mathcal{F}}_i$\;
        Set $\Bar{\mathbf{p}} \gets \Bar{\mathbf{p}} \cup \{p_i\}$, $\Bar{\boldsymbol{\mathcal{F}}} \gets \Bar{\boldsymbol{\mathcal{F}}} \cup \{\mathcal{F}_i\}$\;
        $\boldsymbol{\theta} \gets \text{argmax}\log \prod \mathcal{N} ({\mathcal{F}}(\boldsymbol{f}(\boldsymbol{p}_i))| \boldsymbol{\mu}_{\boldsymbol{\theta}}(\boldsymbol{p}_i), \boldsymbol{K})$; 
        % Set $\Bar{\mathbf{f}} \gets \Bar{\mathbf{f}} \cup \{ q_i - {\mu}_{\boldsymbol{\theta}}(\boldsymbol{p})\}$\;
        Re-estimate GP\;
    }
  }
 \Return Top probe pose with normal direction to $A$\;
 \caption{Bayesian Optimization for Robotized Probe Normalization}
 \label{alg:BO}
\end{algorithm}
\subsection{Objective function of BO}
% {\color{red} Abhishek: Need to rewrite this. It should cover this whole column and beyond 25\%}
When a constant force ${f}^d$ is exerted along the probe axis, it generates a reaction force ${f}_z$ along the normal of the scanning region to balance the applied force ${f}^d$. Consequently, the reaction forces in two mutually perpendicular directions are generated in response to the change in the relative orientation of probe and unknown normal. The concept of reaction force is well-defined in theoretical mechanics, which suggests that if an external force (${f}^d$) is exerted in the direction perpendicular to normal of contact surface, then the corresponding reaction forces in the remaining two orthogonal directions must be zero. Based on this principle, we formulated the objective function of BO (${\mathcal{F}}$) to maximize the reaction force ${f}_z$ and minimize the reaction forces along the other two axis. The mathematical form of ${\mathcal{F}}$ is given as:
\begin{equation} \label{eq:objective_fun}
{\mathcal{F}} = \frac{f_z}{f_x f_y + \lambda ||\boldsymbol{f}||_2 + \epsilon} 
\end{equation}
\begin{figure}[ht]
	\centering
	%trim={L,B,R,T}
	\includegraphics[trim=0cm 4cm 13.5cm 0cm,clip,width=0.7\linewidth]{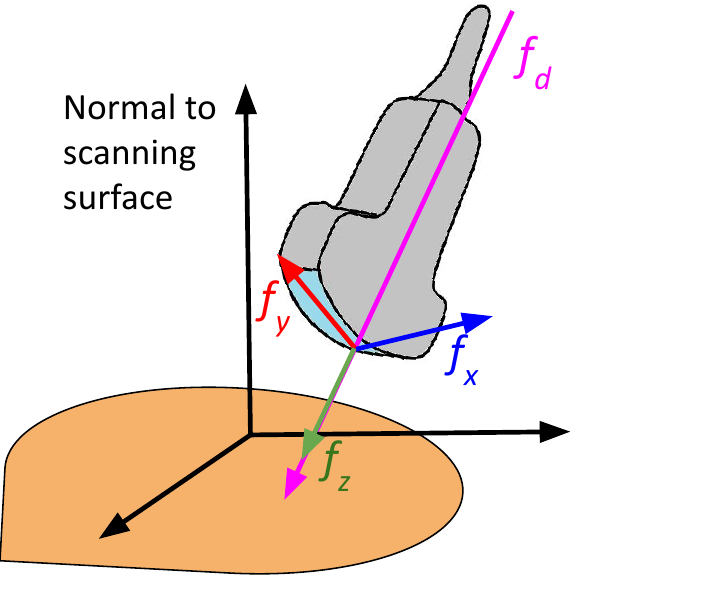}
	\caption{Ultrasound probe generates reactive forces $[f_x, f_y, f_z]$ when a desired force $f_d$ is applied along its axis at a specific orientation to the scanning surface}
	\label{fig:real_setup}
\end{figure}
where $\epsilon$ is a constant term added in the denominator to avoid the division by zero. $\lambda||\boldsymbol{f}||_2$ represents the regularization term, which penalizes the noise in the force sensor by adding a positive value to the denominator of the objective function. The objective function with regularization term can help to stabilize the objective function values and reduce its noisy nature. The regularization term penalizes the reactive force vector, which will prevent the function from becoming too sensitive to small fluctuations in the input, usually because of noisy sensor data. This term would be particularly important in applications where the sensor data in objective function is susceptible to noise or other sources of variability. The $\lambda$ in regularization term is termed as penalty and is usually a positive constant value. The $||\boldsymbol{f}||_2$ denotes the $L2$-norm (also known as euclidean norm) of the reactive force vector $\boldsymbol{f}$, which is given by:
\begin{equation}
||\boldsymbol{f}||_2 = \sqrt{f_x^2 + f_y^2 + f_z^2}
\end{equation}

The choice of $\lambda$ value is critical in determining the strength of the regularization term.  A small value of $\lambda$ would result in a weak regularization and high noise in objective function measurements, while a large value of $\lambda$ will result in low noise, but a large number of samples will be required for converging to small errors. In the current approach, we have chosen $\lambda$ using the iterative selection method, which involves solving the optimization problem for a range of values of $\lambda$ and selecting the value as per performance requirement. 

\subsection{Acquisition function of BO}
To determine the next pose of a probe to sample from during BO search, an acquisition function is used. Given the stochastic nature of the unknown function estimates, we used Expected Improvement (EI) acquisition function for optimizing the next probe pose. The EI function expresses the expected gain in performance if the next probe pose is selected based on the highest predicted value. The improvement is evaluated based on the difference between the highest predicted value and the current best value obtained so far. This allows the acquisition function to balance the exploration of new regions with the exploitation of the current best results.

The EI function takes into account the posterior mean $\boldsymbol{\mu}_{\bar{\boldsymbol{\mathcal{F}}}}(\boldsymbol{p})$ and variance $\boldsymbol{\sigma}_{\bar{\boldsymbol{\mathcal{F}}}}^2(p)$ of the GP model to evaluate the potential improvement of exploring with a given pose.  The EI function can be written as a combination of the cumulative distribution function (CDF) and probability density function (PDF) of the Gaussian distribution. Specifically, it is given by the formula:
\begin{equation} \label{eq:ei}
    \resizebox{0.9\hsize}{!}{$EI(\boldsymbol{p}) = 
    \begin{cases}
    (\boldsymbol{\mu}_{\bar{\boldsymbol{\mathcal{F}}}}(\boldsymbol{p}) - \boldsymbol{\mathcal{F}}^+ - \xi)\boldsymbol{\Phi}(\boldsymbol{Z}) +\boldsymbol{\sigma}_{\bar{\boldsymbol{\mathcal{F}}}}^2(\boldsymbol{p})\boldsymbol{\phi}(\boldsymbol{Z}) & \text{if } \boldsymbol{\sigma}_{\bar{\boldsymbol{\mathcal{F}}}}^2(\boldsymbol{p}) > 0\\
    0              & \text{if } \boldsymbol{\sigma}_{\bar{\boldsymbol{\mathcal{F}}}}^2(\boldsymbol{p}) = 0
    \end{cases}$}
\end{equation}

where $\boldsymbol{\Phi}$ and $\boldsymbol{\phi}$ are the CDF and PDF of the standard normal distribution, respectively. $Z$ is a standard normal variable that is calculated based on the posterior mean and variance of the GP model. $\boldsymbol{\mathcal{F}}^+$ is the best function value found so far, and $\xi$ is a parameter that controls the trade-off between exploration and exploitation.

\section{Results}
\begin{table*}[ht]
  \centering
  % \vspace{-2mm}
    \caption{Absolute angular error between ground truth and predicted values of normal for out-plane and in-plane rotation, respectively, of \textbf{linear probe} for different search limits [min, max]. Each value represents the Mean$\pm$S.D. for 5 runs of BO}
    \resizebox{\linewidth}{!}{\begin{tabular}{cccccccccccccccc}
    \toprule
    \multirow{2}{*}{\parbox{1cm}{\centering\textbf{Phantom surface}}} &  \multicolumn{7}{c}{\textbf{Out-plane rotation (deg.)}} & \multicolumn{7}{c}{\textbf{In-plane rotation (deg.)}} & \multirow{2}{*}{\parbox{1cm}{\centering\textbf{Avg.}}}\\
    \cmidrule(lr){2-8}\cmidrule(lr){9-15}
    % ~ & ~ & \multicolumn{4}{c}{\textbf{Sum of quality difference of $n$ points}} & {\textbf{Top}} & \textbf{ZNCC} & \multicolumn{4}{c}{\textbf{Sum of quality difference of $n$ points}} & {\textbf{Top}} & \textbf{ZNCC} \\
    ~ & $[-5,5]$ & $[-10,5]$ & $[-5,10]$ & $[-10,-10]$ & $[-15,5]$ & $[-5,15]$ & $[-15,15]$ & $[-5,5]$ & $[-10,5]$ & $[-5,10]$ & $[-10,-10]$ & $[-15,5]$ & $[-5,15]$ & $[-15,15]$ & ~\\
    \midrule
    \multirow{2}{*}{\parbox{1cm}{\centering\textbf{Planar}}} & $1.25$ & $1.57$ & $2.77$ & $3.21$ & $3.36$ & $2.93$ & $3.37$ & $1.11$ & $1.65$ & $2.32$ & $1.93$ & $2.13$ & $2.29$ & $3.15$ & $2.36$\\
    ~ & $\pm0.4$ & $\pm0.8$ & $\pm1.1$ & $\pm0.3$ & $\pm1.3$ & $\pm0.8$ & $\pm1.1$ & $\pm0.4$ & $\pm0.2$ & $\pm0.5$ & $\pm1.2$ & $\pm0.8$ & $\pm1.2$ & $\pm0.5$ & $\pm0.4$\\
    % \cmidrule(lr){1-1}\cmidrule(lr){2-8}\cmidrule(lr){9-15}\cmidrule(lr){16-16}
    \midrule
    \multirow{2}{*}{\parbox{1cm}{\centering\textbf{Tilted}}} & $1.45$ & $1.72$ & $2.19$ & $2.20$ & $2.75$ & $3.5$ & $3.80$ & $1.23$ & $1.82$ & $2.21$ & $2.76$ & $3.23$ & $3.90$ & $4.61$ & $2.67$\\
    ~ & $\pm0.2$ & $\pm0.3$ & $\pm0.7$ & $\pm0.4$ & $\pm0.6$ & $\pm0.9$ & $\pm1.7$ & $\pm0.4$ & $\pm1.1$ & $\pm0.7$ & $\pm0.3$ & $\pm0.9$ & $\pm0.7$ & $\pm0.1$ & $\pm0.6$\\
    % \cmidrule(lr){1-1}\cmidrule(lr){2-8}\cmidrule(lr){9-15}\cmidrule(lr){16-16}
    \midrule
    \multirow{2}{*}{\parbox{1cm}{\centering\textbf{Rough}}} & $1.37$ & $2.21$ & $2.38$ & $2.56$ & $2.22$ & $2.71$ & $3.32$ & $1.61$ & $1.72$ & $1.13$ & $2.11$ & $2.05$ & $2.72$ & $3.12$ & $2.23$\\
    ~ & $\pm0.3$ & $\pm1.1$ & $\pm1.4$ & $\pm0.5$ & $\pm1.8$ & $\pm0.9$ & $\pm1.7$ & $\pm0.4$ & $\pm1.5$ & $\pm0.9$ & $\pm1.3$ & $\pm0.9$ & $\pm1.2$ & $\pm0.7$ & $\pm1.0$\\
    % \cmidrule(lr){1-1}\cmidrule(lr){2-8}\cmidrule(lr){9-15}\cmidrule(lr){16-16}
    \midrule
    \multirow{2}{*}{\parbox{1cm}{\centering\textbf{Avg.}}}& $1.35$ & $1.83$ & $2.44$ & $2.65$ & $2.77$ & $3.04$ & $3.49$ & $1.31$ & $1.73$ & $1.88$ & $2.26$ & $2.47$ & $2.97$ & $3.63$ & $\boldsymbol{2.44}$\\
    ~ & $\pm0.3$ & $\pm0.7$ & $\pm1.1$ & $\pm0.4$ & $\pm1.2$ & $\pm0.86$ & $\pm1.5$ & $\pm0.4$ & $\pm1.2$ & $\pm0.7$ & $\pm0.93$ & $\pm0.9$ & $\pm0.9$ & $\pm0.7$ & $\boldsymbol{\pm0.7}$\\   
    \bottomrule
  \end{tabular}}
  \label{tab:bo_phantoms}
\end{table*}

\subsection{Implementation details}
We validated the proposed framework in both real and simulated environment. 
The real environment consists of a Sawyer 7-DOF robotic arm (Rethink Robotics, Germany) with an ultrasound probe attached to its end-effector using a customized gripper, as shown in Fig. \ref{fig:real_setup}. 
\begin{figure}[ht]
	\centering
	%trim={L,B,R,T}
	\includegraphics[trim=0cm 4cm 7cm 0cm,clip,width=\linewidth]{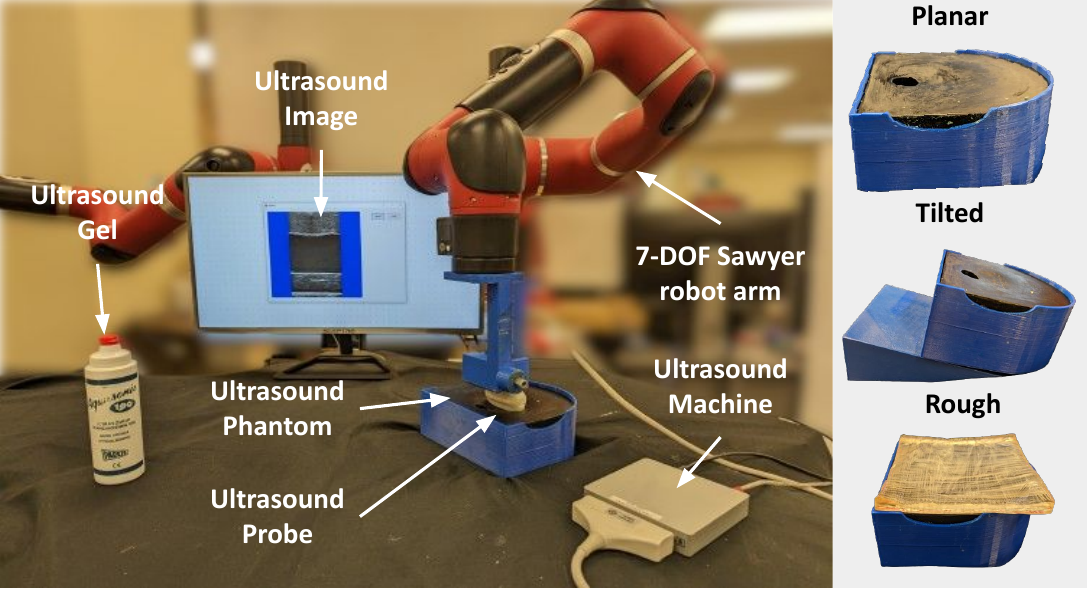}
	\caption{Experimental setup of RUS with urinary bladder phantom having three different scanning surfaces.}
	\label{fig:real_setup}
\end{figure}
This arm has a wrist-mounted 6-axis force sensor, which is used for real-time contact force measurement during the scanning. The two ultrasound probes are used, one is Linear L12-5N40-M3 and the other is a Convex MC10-5R10S-3 probe, which is connected to a Telemed ultrasound machine (Telemed Medical Systems, Italy). The scanning is conducted on urinary bladder phantom (YourDesignMedical, USA). The phantom surface is flat, however it has been modified to represent tilted and rough surface to evaluate the robustness of the framework for different human physiological conditions. The rough surface has been made by using a Ballistics gel (Clear Ballistics, USA) layer.

\begin{figure}[!ht]
	\centering
	%trim={L,B,R,T}
	\includegraphics[trim=0cm 6cm 9cm 0cm,clip,width=\linewidth]{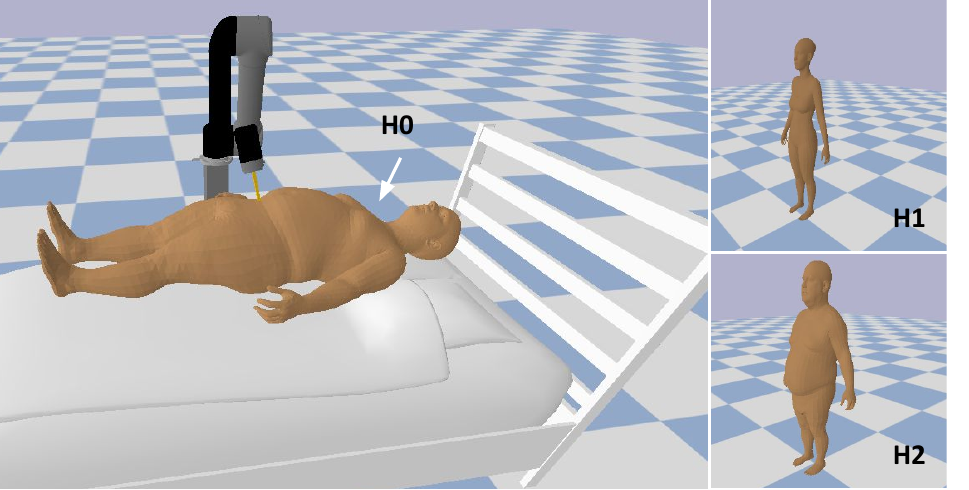}
	\caption{Simulation environment of robotic ultrasound system with 3D human mesh models denoted as H0, H1 and H2.}
	\label{fig:sim_setup}
\end{figure}

For simulation experiments, we built a physics-based simulation framework using the open-source Pybullet physics engine \cite{coumans2016pybullet}, as shown in Fig. \ref{fig:sim_setup}. 
Further, we imported 3D mesh models of the human body using SMPL-X \cite{pavlakos2019expressive}, which represent the actual physiology of real humans. These models were developed for capturing human posture using RGB and depth images \cite{pavlakos2019expressive}. Our simulation platform will help us to do thousands of experiments safely, which would be difficult to conduct with real-world robotic system and human subjects. For both simulation and real experiments, we used the value of $\epsilon=1.0$, $\xi=0.45$ and $\lambda=0.3$ for BO. The robot is under hybrid-position control \cite{raina2021comprehensive}, where z-axis of robot is under force control.

% \begin{figure*}[!ht]
% 	\centering
% 	%trim={L,B,R,T}
% 	% \includegraphics[trim=0cm 3cm 0cm 0cm,clip,width=\linewidth]{figs/overview9}
% 	\includegraphics[trim=0cm 5cm 0cm 0cm,clip,width=\linewidth]{figs/rotation_Series}
% 	\caption{Impact of out-plane and in-plane ultrasound probe rotations on ultrasound image quality. The upper row shows the probe rotations and bottom row shows the corresponding ultrasound images}
% 	\label{fig:robot_us}
% \end{figure*}
\subsection{Validation metrics}
To verify the accuracy of the obtained normal direction vector in the simulated environment, we utilized a mesh of the body being scanned. We compared the predicted normal to the true normal of the scanning region obtained from the mesh geometry. The mesh of the body returns the triangles (faces) and the coordinates of the connecting points that connect the triangles. We used a ray-mesh intersection method to determine the intersection between the probe point and mesh. This method utilizes a Bounding Volume Hierarchy (BVH) tree to quickly eliminate parts of the mesh that the ray does not intersect with. After identifying potential intersecting triangles, the method performs more precise intersection tests using Gilbert–Johnson–Keerthi (GJK) algorithm. This is implemented using Trimesh \cite{dawsontrimesh} python library. For experiments with real robots, point cloud data has been utilized to estimate the normal vector. This was accomplished using the PCL Python library and subsequently corrected based on known geometry. The true normal vector for planar and rough phantoms lies along the vertical z-axis in the robot base frame. Similarly, the normal vector for tilted phantoms was estimated using a geometric method, as the tilt angle was known beforehand.

\subsection{Validation on urinary-bladder phantom}
To evaluate the algorithm's robustness for clinical ultrasound applications, we conducted a set of BO experiments on phantoms with flat, tilted, and rough surfaces. We performed these experiments for seven different limits ([min., max.]) of the search space. Each BO run was initiated with a probe positioned at the center of the phantom and random initial orientation. The resulting angular differences between the estimated normal and the actual normal are presented in Table \ref{tab:bo_phantoms}. Each value represents the mean and standard deviation across five BO runs. In total, we conducted $210$ BO runs for results presented in Table \ref{tab:bo_phantoms}.

From Table \ref{tab:bo_phantoms}, it can be observed that the average angular error across all runs is $2.44\pm0.7^\circ$, which is lower than the error reported for human sonographers in \cite{jiang2020automatic} as $3.21\pm1.7^\circ$. Additionally, in $165$ out of $210$ runs (i.e., $78.6\%$ of the runs), the error was found to be less than $3.0^\circ$. It is noteworthy that the error values remained consistent across different search space limits for both in-plane and out-plane rotations. When grouped by rotation direction, the max. average error for out-plane and in-plane rotation is found to be close to each other (difference less than $0.5^\circ$). These results indicate that the rotation direction does not have a significant impact on the normal identification. However, the maximum error of in-plane rotation ($4.71^\circ$) is less than that of out-plane rotation ($5.5^\circ$). The variation in the optimization performances for in-plane and out-plane rotations can be attributed to the linear probe's structure, for which the length is significantly greater than the width. Thus, it can be concluded that the in-plane optimization results are more accurate in comparison to out-plane optimization. There are a few sub-optimal cases (error $> 3^\circ$) that occurred in fewer than $22.4\%$ of the runs. This typically occurs when the search space limit is high, which can be attributed to the decrease in the performance of BO for larger search spaces. The effect of this error in the normal identification can be illustrated through Fig. \ref{fig:teaser}. As the rotation angle moves away from normal ($>3^\circ$) orientation, the image shows artifacts like edge or structure shadowing.

We conducted additional experiments using a convex probe for scanning. For curved probe, the significant difference in the ultrasound image is only observed after $15^\circ$. Therefore, we tested it for larger search space limits, which are presented in Table \ref{tab:error_curved}. 
We performed 30 BO runs, and the results showed an average angular difference of $2.31\pm0.7^\circ$ across all runs. In 23 out of 30 runs (i.e., 76.7\% of runs), the error was less than $3.0^\circ$. Notably, the convex probe's error was even lower than that of the linear probe,
\begin{table}[ht]
  \centering
  % \vspace{-2mm}
    \caption{Absolute angular error for out-plane and in-plane rotation of \textbf{convex probe}. Each value represents the Mean$\pm$S.D. for five runs of BO}
    \resizebox{\linewidth}{!}{\begin{tabular}{cccccc}
    \toprule
    \multirow{2}{*}{\parbox{1cm}{\centering\textbf{Phantom surface}}} &  \multicolumn{2}{c}{\textbf{Out-plane rotation (deg.)}} & \multicolumn{2}{c}{\textbf{In-plane rotation (deg.)}} & \multirow{2}{*}{\parbox{1cm}{\centering\textbf{Avg.}}}\\
    \cmidrule(lr){2-3}\cmidrule(lr){4-5}
    ~ &  $[-15,-15]$ & $[-20,20]$ &  $[-15,-15]$ & $[-20,20]$ & ~\\
    \midrule
    {\textbf{Planar}} & $2.11\pm0.8$ & $2.45 \pm 0.3$ & $2.01 \pm 0.6$ & $1.85\pm1.1$ & $2.11\pm0.7$\\
    % \hline
    {\textbf{Tilted}} & $1.99\pm1.0$ & $2.23 \pm0.5$ & $2.12 \pm0.9$ & $2.43\pm0.5$ & $2.19\pm0.7$\\ 
    % \hline
    {\textbf{Rough}} & $2.33\pm1.1$ & $2.89 \pm0.8$ & $2.45 \pm0.3$ & $2.80\pm0.9$ & $2.62\pm0.8$\\ 
    \midrule
    {\textbf{Avg.}} & $2.14\pm1.0$ & $2.52 \pm0.5$ & $2.19 \pm0.6$ & $2.36\pm0.9$ & $\boldsymbol{2.31\pm0.7}$\\ 
    \bottomrule
  \end{tabular}}
  \label{tab:error_curved}
\end{table}
 which is attributed to the former's smaller tip, resulting in smoother force variations.

\subsection{Validation on 3D mesh human models}
\begin{figure}[!ht]
	\centering
	%trim={L,B,R,T}
	\includegraphics[trim=0cm 0.5cm 0cm 0.5cm,clip,width=0.9\linewidth]{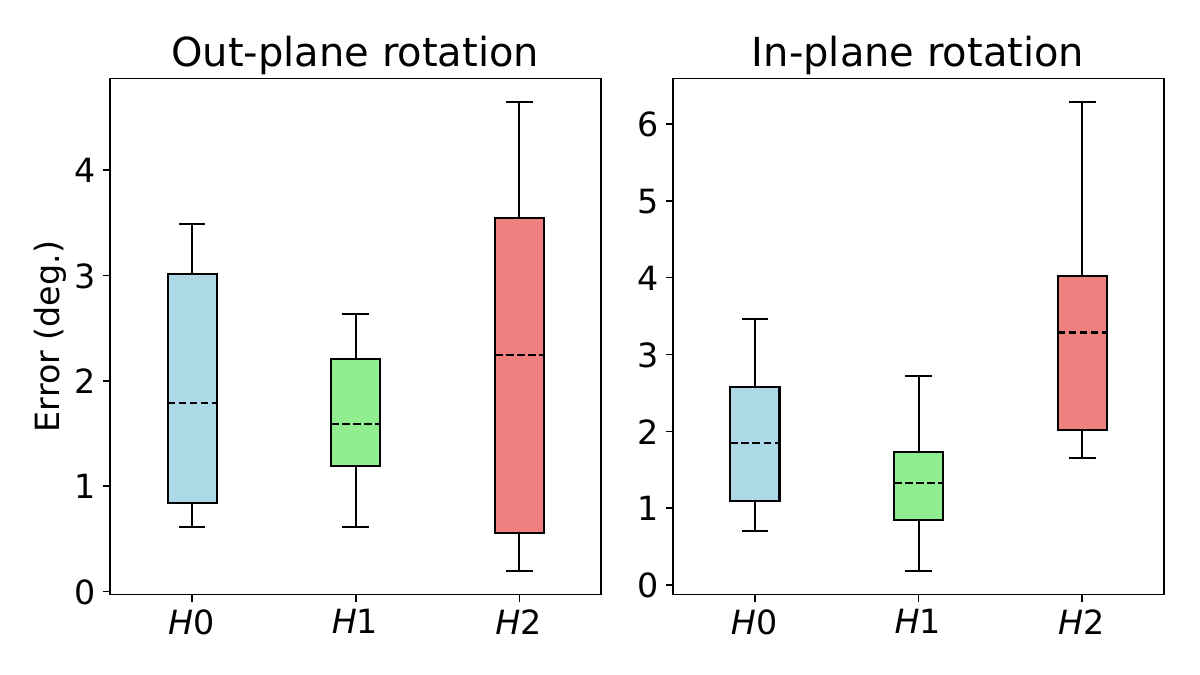}
	\caption{Absolute angular error between ground truth normal and estimated normal for 3D mesh models of human}
	\label{fig:box-plots}
\end{figure}

In order to provide comprehensive validation, the proposed framework was also evaluated on three 3D human mesh models in simulation. The absolute angular differences for the out-plane and in-plane rotations are presented in Fig. \ref{fig:box-plots}. The search space limit for both out-plane and in-plane experiments was set to $[-15,15]$, and 10 runs of BO were conducted for each human model. It was observed that the average error for H0 and H1 was less than $3.0^\circ$, indicating the robustness of the method for complex human physiology. The average error for H3 was relatively high, which can be attributed to the high irregularity of the body surface. Across all $60$ BO runs (in-plane and out-plane), more than $75\%$ of the runs reported an error less than $3.0^\circ$. It is important to note that simulation study has been limited to small contact point on the mesh bodies due to their rigid surfaces. Despite this limitation, it is noteworthy that simulation studies offer a reliable means of verifying the proposed approach for determining the normal direction of the probe while interacting with different 3D human models.
% \begin{table}[ht]
%   \centering
%   % \vspace{-2mm}
%     \caption{Performance of BO on Human Models}
%     \resizebox{\linewidth}{!}{\begin{tabular}{ccccc}
%     \toprule
%     \centering\textbf{Human Model} & \textbf{Out-plane rotation} & \textbf{In-plane rotation}\\
%     % \cmidrule(lr){2-8}\cmidrule(lr){9-15}
%     % ~ & ~ & \multicolumn{4}{c}{\textbf{Sum of quality difference of $n$ points}} & {\textbf{Top}} & \textbf{ZNCC} & \multicolumn{4}{c}{\textbf{Sum of quality difference of $n$ points}} & {\textbf{Top}} & \textbf{ZNCC} \\
%     \midrule
%     \multirow{1}{*}{\textbf{Solid Human}} ~ $$ & $$ &\\
%     \bottomrule
%   \end{tabular}}
%   \label{tab:lambda_val}
% \end{table}
\vspace{-5mm}
\subsection{Analysis of BO objective function}
The proposed framework incorporates a regularisation term in the objective function of BO (eq. \ref{eq:objective_fun}). Hence, we conducted an analysis to examine the effect of this term by varying the penalty parameter, $\lambda$. The outcomes of this analysis are presented in Table \ref{tab:lambda_val}, which are generated using linear probe for scanning different phantom surfaces within the search space of $[-15,15]$. It is apparent from the results that the absence of a regularization term (i.e., $\lambda = 0.0$) results in high error rates ($>3.7^\circ$). However, the addition of the regularization term results in a decrease in error rates. Although further decrease in error rates can be achieved with larger values of $\lambda$, more samples will be required to converge to the low error rates. This is because BO will need to sample more points to attain the global maxima among the highly correlated function maps. Although determining the appropriate value of $\lambda$ requires extensive experimentation, it significantly affects the performance of the proposed framework and plays a key role in enhancing the accuracy of normal identification.
\begin{table}[ht]
  \centering
  % \vspace{-2mm}
    \caption{Performance of BO for different regularization in the objective function}
    \resizebox{\linewidth}{!}{\begin{tabular}{ccccc}
    \toprule
    \centering\textbf{Phantom} & $\lambda$ & \textbf{Out-plane rotation} & \textbf{In-plane rotation} &\textbf{Sampled points}\\
    % \cmidrule(lr){2-8}\cmidrule(lr){9-15}
    % ~ & ~ & \multicolumn{4}{c}{\textbf{Sum of quality difference of $n$ points}} & {\textbf{Top}} & \textbf{ZNCC} & \multicolumn{4}{c}{\textbf{Sum of quality difference of $n$ points}} & {\textbf{Top}} & \textbf{ZNCC} \\
    \midrule
    \multirow{4}{*}{\textbf{Planar}} & $0.0$ & $3.71\pm 0.7$ & $4.85\pm 0.5$ & $50$\\ 
    ~ & $0.3$ & $3.37\pm1.1$ & $3.15\pm0.5$ & $100$\\  
    ~ & $0.5$ & $1.95\pm0.3$ & $2.06\pm0.7$ & $170$\\  
    ~ & $10$ & $1.21\pm 0.2$& $1.54\pm0.3$ & $250$\\
    % \cmidrule(lr){1-1}\cmidrule(lr){2-8}\cmidrule(lr){9-15}\cmidrule(lr){16-16}
    \bottomrule
    \multirow{4}{*}{\textbf{Tilted}} & $0.0$ & $5.14 \pm1.1$ & $4.76 \pm0.9$ & $57$\\  
    ~ & $0.3$ & $3.81\pm1.7$ & $4.61\pm0.1$ & $95$\\  
    ~ & $0.5$ & $2.34\pm0.4$ & $2.17\pm0.7$ &$182$\\  
    ~ & $10$ & $1.43\pm0.2$ & $1.21\pm0.3$ &$271$\\
    % \cmidrule(lr){1-1}\cmidrule(lr){2-8}\cmidrule(lr){9-15}\cmidrule(lr){16-16}
    \bottomrule
        \multirow{4}{*}{\textbf{Rough}} & $0.0$ & $4.13\pm0.5$ & $3.67\pm0.6$ & $63$\\
        ~ & $0.3$ & $3.32\pm1.7$ & $3.12\pm0.7$ & $121$\\  
    ~ & $0.5$ & $2.05\pm0.2$ & $1.93\pm0.6$ & $168$\\  
    ~ & $10$ & $1.12\pm0.3$ & $1.32\pm0.2$ & $263$\\  
    \bottomrule
  \end{tabular}}
  \label{tab:lambda_val}
\end{table}

% - For every surface, we need to do 10 or 20 runs and find errors in normal, mean and std dev of error\\
% - Represent results in table first\\
% \textbf{Validation on different human mesh models}\\
% - Similar results as above\\
% {\color{red}{DEADLINE: 11-FEB}}\\

% \textbf{Real-robot studies}\\
% - Image quality analysis
% - Phantom in different orientations - same results as simulation\\
% \textbf{COMPARISON with other methods ????}
% \vspace{-5mm}
\section{Conclusion}
In this paper, we propose a novel Bayesian Optimization (BO) based method for determining the normal direction of the probe to the scanning region during robotic ultrasound procedures, considering both in-plane and out-plane rotations of the robotized probe. Our approach introduces a novel objective function for BO that leverages force sensor measurements and the underlying mechanics of contact to guide the identification process. Notably, this method offers significant advantages as it does not rely on a 3D point cloud of the patient's body and extensive exploration for finding normal. To evaluate its performance, we conducted experiments using urinary bladder phantoms and 3D mesh models of humans, employing both linear and convex probes.

However, it is important to note that this study represents a preliminary step towards achieving an efficient estimation of the probe's normal orientation, and further investigation is required. One major limitation of the study is the absence of ultrasound imaging feedback for normal identification. Our future work will explore the joint optimization of force and ultrasound image quality \cite{raina2022slim} using multi-objective Bayesian optimization. Another limitation of the work is the adjustment of probe orientation at a single point. In the future, we will explore the normalization of the probe during continuous scanning. Furthermore, we aim to extend our framework to optimize varying probe orientations and desired forces, thereby increasing its clinical applicability for scanning complex physiology and different anatomical structures under skin, bones, or muscles. To achieve this, we will explore the integration of our method with the RUS framework presented in our previous works \cite{raina2023deep, raina2023robot, raina2023robotic}, incorporating the geometrical information of the anatomy being scanned.  
To further enhance the sample efficiency of the optimization process, we will explore leveraging the domain expertise in BO. For instance, a potential solution involves using a prior in BO as suggested in \cite{raina2023robotic, zhu2022automated}, which can be modeled using expert's demonstrations.
Finally, in order to validate the proposed framework for clinical procedures, it will be necessary to conduct assessments on humans. In future, we intend to conduct experiments on human subjects using our robotic ultrasound system, proposed in our earlier works \cite{raina2021comprehensive, chandrashekhara2022robotic}. These experiments will provide insights into the clinical applicability of our method, establishing its effectiveness in real-world medical settings.
\bibliography{references} 
\bibliographystyle{ieeetr}
% \end{IEEEbiography}
% % insert where needed to balance the two columns on the last page with
% %\newpage
%\vfill
\end{document}